\documentclass{article} %
\usepackage{iclr2027_conference,times}

\usepackage{amsmath,amsfonts,bm}

\def\eqref#1{equation~\ref{#1}}

\def\1{\bm{1}}

\DeclareMathAlphabet{\mathsfit}{\encodingdefault}{\sfdefault}{m}{sl}
\SetMathAlphabet{\mathsfit}{bold}{\encodingdefault}{\sfdefault}{bx}{n}

\usepackage{url}
\usepackage{graphicx}
\usepackage{subcaption}
\usepackage{wrapfig}
\usepackage{booktabs}
\usepackage{xcolor}
\usepackage{colortbl}

\usepackage{enumitem}
\newcommand{\dg}[1]{\textcolor{black!45}{#1}}
\newcommand{\up}[1]{{\scriptsize\color{green!55!black}($\uparrow$#1)}}
\newcommand{\dn}[1]{{\scriptsize\color{black!45}($\downarrow$#1)}}  %
\usepackage{multirow}
\graphicspath{{figures/}}

\usepackage[hidelinks]{hyperref}

\title{PACT-VLA: Speculative Action-Chunk Verification with Pre-Commit Observations}

\author{Yixiang Zhu$^{1,2}$\enspace Ruilong Wu$^{1}$\enspace Yonghao Chen$^{1}$\enspace Jiaxiang Zou$^{1}$\enspace Yusong Hu$^{2}$\enspace Xinyu Chen$^{1}$ \\
{\normalfont $^{1}$The Hong Kong University of Science and Technology (Guangzhou)\quad $^{2}$One Robotics}
}
\iclrfinalcopy

\begin{document}

\maketitle
\lhead{}\renewcommand{\headrulewidth}{0pt}

\begin{abstract}
Vision-language-action (VLA) policies are commonly deployed with asynchronous inference in real-world settings, where the next action chunk is generated while the robot is still executing the previous one. By the time the new chunk is committed, the robot and scene have changed, so the chunk is conditioned on a stale observation and task success drops sharply as this inference delay grows. Generating multiple candidate chunks could let newer observations inform the final choice, but existing multi-candidate methods either rank candidates
with the stale inference-start observation or repeatedly switch between
candidates during execution, which can make the executed action sequence
discontinuous. Our key insight is that inference delay can be used as decision time. While
candidate generation is still underway, candidate selection can reuse the VLA
itself to choose the complete action chunk that best fits a later observation. We propose PACT-VLA, a training-free framework that generates candidates from the inference-start observation and selects one using this pre-commit observation. Its scoring function, FlowRes, measures the residual between each candidate's implied flow direction and the velocity predicted by the frozen policy under the pre-commit observation, requiring no separate verifier. A pipelined implementation overlaps observation processing with candidate generation and batches scoring, so selection adds no control-step delay on the evaluated hardware. At a four-step delay, PACT-VLA improves average success by 15.0 percentage points over RTC on Kinetix and by 5.2 points over single-candidate asynchronous execution with GR00T N1.7 on LIBERO, and it further improves existing delay-robust methods. On three real-robot tasks, it raises average success from 60\% to 82\% over naive asynchronous execution with the same frozen $\pi_{0.5}$ checkpoint.
\end{abstract}

\section{Introduction}
\label{sec:introduction}

Vision-language-action (VLA) policies~\citep{brohan2023rt,kim2024openvla,wen2024diffusion} increasingly predict action chunks rather than individual actions~\citep{li2024cogact,liu2025rdt,black2024pi_0,team2024octo}, allowing several control steps to be executed from a single policy inference and preserving temporal coherence across consecutive actions~\citep{black2026real,zhao2023learning,kim2025fine}. However, action-chunk generation is not instantaneous. While inference is running, the robot continues to act and the surrounding scene may change. Consequently, by the time a newly generated chunk is ready for execution, the observation from which it was generated may already be stale, and the chunk may no longer fit the scene~\citep{black2026real,tang2025vlash,agouzoul2026understanding}.
The challenge is that action generation begins before later changes are observed.

Existing approaches address this challenge in different ways. Future-state methods use information available at inference start to
anticipate the conditions under which the next action chunk will be executed~\citep{tang2025vlash}.
However, these methods determine the next action chunk before changes during
the current inference interval are observed, so the conditions at execution
may differ from what was anticipated. A natural alternative is to postpone
this decision by keeping multiple candidate action chunks available until more
information about the upcoming execution conditions becomes available. However, some test-time
sampling methods still rank these candidates using the inference-start
observation~\citep{kwok2025robomonkey,jang2026verifier}, so the final choice
remains fixed before later changes are observed. 

DreamChunk~\citep{chen2026dream}
goes further by using a learned world model to predict a trajectory for each
candidate. As new observations arrive during execution, it compares them with
these predicted trajectories and repeatedly reselects which candidate to
follow. 
This allows the choice to respond to newly observed changes, but keeping the
decision open during execution can cause repeated candidate switching, making
the executed action sequence discontinuous, which can hurt task performance. However, sensing
continues while the next chunk is being generated, so an observation arriving
during inference can already be available before commitment. This leaves open
whether such an observation can be used to finalize one complete action chunk
before commitment, without switching between candidates during execution.

Our key insight is that inference delay can be used as decision time. While
candidate generation is still underway, candidate selection can reuse the VLA
itself to choose the complete action chunk that best fits a later observation.
We propose PACT-VLA, which realizes this principle by generating a fixed
candidate pool from the inference-start observation and selecting one candidate
using a pre-commit observation received shortly before execution. To choose among the candidates, PACT-VLA uses the original
flow-based VLA policy~\citep{lipman2022flow,black2024pi_0} to evaluate each
candidate under the pre-commit observation. We introduce FlowRes, which
measures the mismatch between the flow direction implied by each candidate
and the direction predicted by the frozen policy under this observation.
The candidate with the lowest FlowRes score is selected, and the controller
executes actions from the selected chunk until the next planned chunk switch.

Using a pre-commit observation leaves only a short interval for candidate selection before the planned chunk switch. PACT-VLA therefore processes the pre-commit observation while candidate generation is still running and scores all candidates in a single batch. On the evaluated hardware, the full generation-and-selection cycle remains in the same control-step delay bucket as single-candidate generation. Experiments on Kinetix, LIBERO, and three real-robot tasks demonstrate the benefit of this design. At a four-step delay, PACT-VLA improves average success by 15.0 percentage points over RTC on Kinetix and by 5.2 points over single-candidate asynchronous execution with GR00T N1.7 on LIBERO. Across the three real-robot tasks, average success increases from 60\% to 82\% using the same frozen $\pi_{0.5}$ checkpoint.

\paragraph{Contributions.}
Our contributions are as follows.
\begin{itemize}[leftmargin=2.0em]
    \item We propose PACT-VLA, a training-free framework for action selection
    under inference delay. PACT-VLA uses observations arriving during inference
    to select one complete action chunk before commitment, reducing the
    prediction--execution mismatch.
    
    \item We introduce FlowRes, a candidate-verification method that reuses the
    frozen VLA to rank candidates under the pre-commit observation, without
    additional training or a separate verifier.

    \item We design a pipelined implementation that overlaps pre-commit observation processing with candidate generation and batches candidate verification.

    \item Experiments on Kinetix, LIBERO, and real robots show gains across multiple flow-based VLA policies and further improvements when combined with existing delay-robust methods.
\end{itemize}\section{Background and Motivation}
\label{sec:background}

\subsection{Flow-Based Action-Chunk Generation under Inference Delay}
\label{sec:preliminaries}

\paragraph{Action-chunk policies.}
We consider a Vision-Language-Action (VLA) policy $\pi_\theta$. At control
step $t$, the policy receives an observation $o_t$, containing the visual and
robot-state information available at that step, together with a language
instruction $l_t$. The policy predicts an action chunk
\begin{equation}
    A
    =
    [a_0,\ldots,a_{H-1}]
    \in
    \mathbb{R}^{H\times d_a},
    \label{eq:action_chunk}
\end{equation}
where $H$ is the action-chunk length and $d_a$ is the action dimension.
The actions in $A$ are executed sequentially over multiple control steps
\citep{chi2025diffusion}.

We index time by control steps. With an inference delay of $d$ steps,
a prediction started at step $t$ becomes available at step $t+d$.
During this interval, an asynchronous controller continues executing actions
that were committed earlier \citep{shukor2025smolvla}. The new action chunk
is therefore generated from information available at step $t$ but becomes
available only at step $t+d$, by which time the robot or scene may have
changed. Sensing continues during the same interval.

\paragraph{Flow-based action generation.}
Flow-based VLA policies generate action chunks by learning a conditional
velocity field. Given an action chunk $A$, Gaussian noise
$\epsilon\sim\mathcal{N}(0,I)$, and flow time $\tau\in[0,1]$, consider the
linear interpolation \citep{liu2022flow}
\begin{equation}
    x_\tau
    =
    (1-\tau)A+\tau\epsilon,
    \qquad
    u_\tau
    =
    \frac{\partial x_\tau}{\partial \tau}
    =
    \epsilon-A.
    \label{eq:flow_path}
\end{equation}
Here, $x_0=A$ and $x_1=\epsilon$. The policy learns an observation- and
language-conditioned velocity field
\citep{lipman2022flow,lipman2024flow} by minimizing
\begin{equation}
    \mathcal{L}_{\mathrm{FM}}(\theta)
    =
    \mathbb{E}_{\substack{
        (o,l,A)\sim\mathcal{D},\,
        \tau\sim p_\tau,\,
        \epsilon\sim\mathcal{N}(0,I)
    }}
    \left[
        \left\|
            v_\theta(x_\tau,\tau\mid o,l)
            -
            u_\tau
        \right\|_2^2
    \right],
    \label{eq:fm_prelim}
\end{equation}
where $\mathcal{D}$ is the training distribution and $p_\tau$ is the
flow-time sampling distribution.

At inference time, the policy starts from Gaussian noise at $\tau=1$
and integrates the learned velocity field toward an action chunk as
$\tau$ decreases to $0$. Different noise samples under the same conditioning
can produce multiple plausible action chunks, providing a natural source of
candidate diversity at test time. In particular, sampling from the
inference-start information yields a candidate pool
\begin{equation}
    \mathcal{A}_t
    =
    \left\{
        A_t^{(1)},\ldots,A_t^{(N)}
    \right\},
    \label{eq:candidate_pool}
\end{equation}
where $N$ is the number of candidate chunks. All candidates are generated
from the information available when inference begins. Meanwhile, sensing
continues during inference, so later observations may reveal changes in the
robot and scene after candidate generation has begun. These changes are not yet observable when generation begins.

\begin{figure}[t]
    \centering
    \captionsetup[subfigure]{justification=centering}
    \begin{subfigure}[t]{0.317\linewidth}
        \centering
        \includegraphics[width=\linewidth]{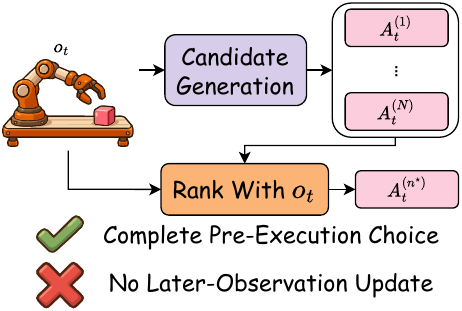}
        \caption{Same-observation ranking}
        \label{fig:motivation_a}
    \end{subfigure}
    \hfill
    \begin{subfigure}[t]{0.342\linewidth}
        \centering
        \includegraphics[width=\linewidth]{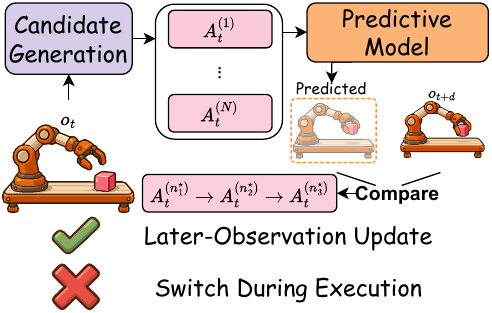}
        \caption{World-model-guided reselection}
        \label{fig:motivation_b}
    \end{subfigure}
    \hfill
    \begin{subfigure}[t]{0.317\linewidth}
        \centering
        \includegraphics[width=\linewidth]{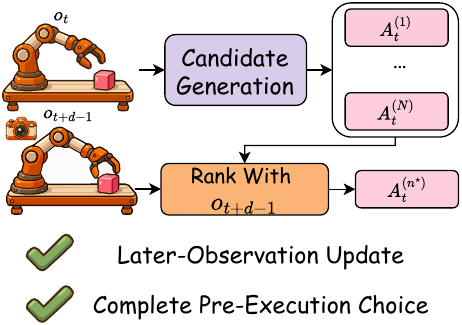}
        \caption{Pre-commit ranking (ours)}
        \label{fig:motivation_c}
    \end{subfigure}
    \caption{
\textbf{Candidate-selection paradigms under delayed execution.}
(a) Candidates are generated and ranked using the same observation $o_t$.
(b) A learned predictive model predicts a trajectory for each candidate and
repeatedly compares these trajectories with later observations during
execution. $n_1^\star$, $n_2^\star$, and $n_3^\star$ denote the candidate indices
selected at successive reselection points and may differ as new observations
arrive.
(c) PACT-VLA uses the original flow policy to rank the fixed candidate pool
under the pre-commit observation $o_{t+d-1}$ before the planned chunk switch.
}
    \label{fig:selection_paradigms}
\end{figure}

\subsection{Action Generation and Selection under Inference Delay}
\label{sec:existing_approaches}
Existing methods address the prediction–execution mismatch in different ways.
One approach is to predict what the robot will be like when the next chunk is
executed. VLASH~\citep{tang2025vlash} predicts the robot state at the expected
execution time and generates the next action chunk based on that prediction.
However, the robot may continue to change during inference, so its state at
execution may differ from the prediction.

A different approach is to keep multiple candidate action chunks available
instead of fixing one chunk immediately. Some test-time sampling methods
generate multiple candidates but rank them using the inference-start
observation $o_t$, as shown in Figure~\ref{fig:selection_paradigms}(a)
\citep{kwok2025robomonkey,jang2026verifier}. Later observations therefore
cannot change which candidate is selected.
DreamChunk~\citep{chen2026dream} goes further by using a learned world model
to predict a trajectory for each candidate. As new observations arrive during
execution, it compares them with the predicted trajectories and repeatedly
reselects which candidate to follow, as shown in
Figure~\ref{fig:selection_paradigms}(b). This allows the choice to respond to
newly observed changes, but keeping the decision open during execution can
cause repeated candidate switching, making the executed action sequence
discontinuous, which can hurt task performance.
Sensing, however, continues while the next chunk is being generated, so a
later observation can already be available before commitment. 
This leaves
open whether the inference interval itself can be used to complete the choice
of one complete action chunk before commitment.

\subsection{Our Insight}
\label{sec:precommit_insight}

Our key insight is that inference delay can be used as a decision-making window. While candidate generation is still ongoing, the selection process can already leverage a later observation, and the VLA itself can determine which complete action chunk best fits that observation before commitment.
PACT-VLA builds on this idea by generating a fixed candidate pool from the
inference-start observation and completing selection with a pre-commit
observation. Section~\ref{sec:method} describes how FlowRes reuses the frozen
VLA to evaluate the candidates and how pipelined execution completes the
selection before the planned chunk switch.\section{Method}
\label{sec:method}

\subsection{Overview}
\label{sec:overview}

\begin{figure}[b]
    \centering
    \includegraphics[width=\linewidth]{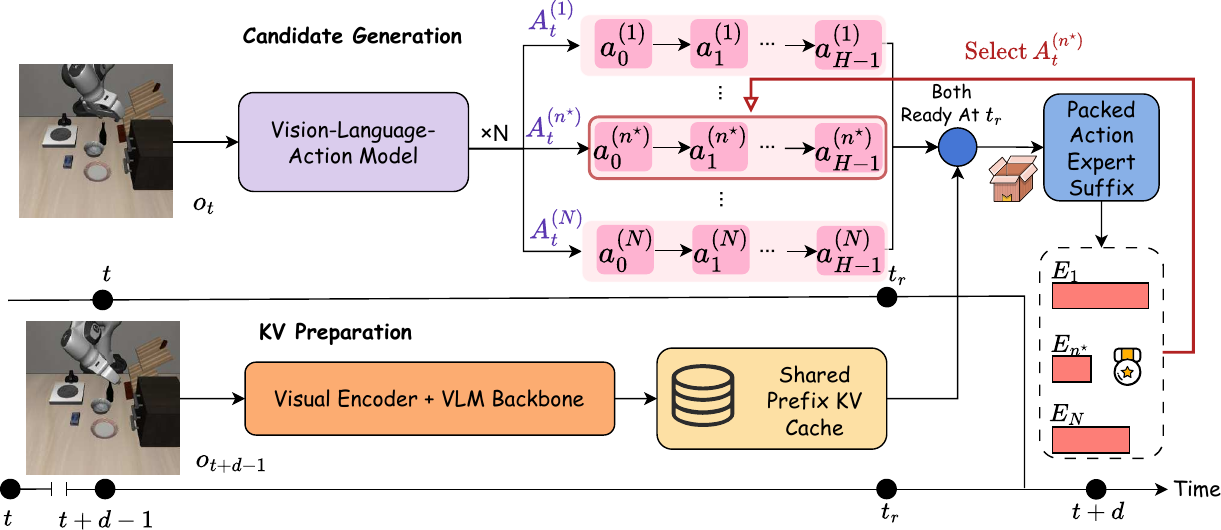}
    \caption{
    \textbf{Overview of PACT-VLA.}
    At control step $t$, PACT-VLA uses the policy's original generation
    procedure to produce a fixed pool of candidate action chunks from the
    information available at inference start. Before the planned chunk
    switch at $t+d$, the pre-commit observation $o_{t+d-1}$ is used to
    evaluate the fixed candidates with FlowRes. The candidate with the
    lowest score is selected, and the controller switches to this chunk
    at $t+d$. The lower branch illustrates how pre-commit observation processing can overlap with candidate generation, while the resulting representation is reused for batched candidate scoring.
    }
    \label{fig:pact_pipeline}
\end{figure}

Figure~\ref{fig:pact_pipeline} shows how PACT-VLA realizes the
separation between candidate generation and final selection introduced
in Section~\ref{sec:precommit_insight}. At control step $t$, the VLA
begins generating $N$ candidate action chunks
$A_t^{(1)},\ldots,A_t^{(N)}$ using its original candidate-generation
procedure and the information available at inference start. These
candidates form a fixed pool and are not modified by observations
received later.

For delayed execution with $d\geq 1$, the controller is scheduled to
switch to the next chunk at step $t+d$. PACT-VLA uses
$o_{t+d-1}$, the observation from one control step before this planned
switch, so that one control-step interval remains for observation
processing and candidate selection. We call $o_{t+d-1}$ the
\emph{pre-commit observation}. The frozen flow policy evaluates the
fixed candidates under this observation and assigns FlowRes scores
$E_1,\ldots,E_N$. PACT-VLA selects the candidate with the lowest
FlowRes score. At $t+d$, the controller switches to the selected chunk
and executes actions from this chunk until the next planned chunk
switch.

Because the pre-commit observation is received close to the planned
switch, its processing is pipelined with the remaining candidate-generation
computation. For VLA backbones with a reusable visual-language prefix,
the visual encoder and VLM backbone process $o_{t+d-1}$ once to prepare
a shared prefix KV cache. Once both the candidate pool and the shared
cache are ready, all candidates are scored together in a single batched
action-expert pass. For flow policies without this module separation,
the same verification is implemented as a batched velocity query.
The following subsections describe FlowRes verification and pipelined execution in detail.

\subsection{FlowRes Candidate Verification}
\label{sec:flowres}

\setlength{\intextsep}{0pt}
\begin{wrapfigure}[27]{r}{0.50\textwidth}
    \centering
    \setlength{\abovecaptionskip}{2pt}
    \includegraphics[width=\linewidth]{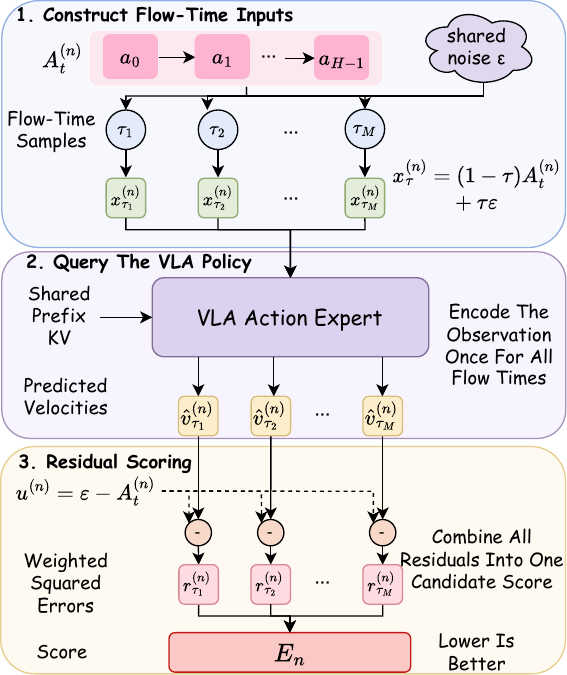}
    \caption{
    \textbf{FlowRes verification of one candidate.}
    PACT-VLA constructs flow-time inputs from the fixed candidate and
    shared verification noise, queries the frozen policy under the
    pre-commit observation, computes the resulting velocity residuals,
    and aggregates them into candidate score $E_n$.
    }
    \label{fig:flowres}
\end{wrapfigure}
\setlength{\intextsep}{12pt plus 2pt minus 2pt}

Given the fixed candidate pool
$\mathcal{A}_t=\{A_t^{(1)},\ldots,A_t^{(N)}\}$,
PACT-VLA evaluates each candidate under the pre-commit observation.
The pointwise velocity residual underlying the flow-matching objective
in Section~\ref{sec:preliminaries} compares the flow direction implied
by an action chunk with the direction predicted by the policy under a
conditioning observation. We use this velocity mismatch as a
policy-consistency signal for candidate verification.  Figure~\ref{fig:flowres}
illustrates the computation for one candidate.

Let $\mathcal{T}=\{\tau_1,\ldots,\tau_M\}$ denote the $M$
verification flow times. For verification, PACT-VLA draws one noise
tensor
$\epsilon^{\mathrm{ver}}\sim\mathcal{N}(0,I)$ independently of the
randomness used to generate the candidates. The same
$\epsilon^{\mathrm{ver}}$ is shared across all candidates and
verification flow times \citep{glasserman1992some}. Sharing the same
verification noise keeps the noise realization fixed across candidate
comparisons. Using the same noise across flow times also places the
verification points of each candidate on a single interpolation path.
For candidate $A_t^{(n)}$ and verification flow time $\tau_i$, we
construct
\begin{equation}
\begin{aligned}
    x_{\tau_i}^{(n)}
    &=
    (1-\tau_i)A_t^{(n)}
    +
    \tau_i\epsilon^{\mathrm{ver}},
    \\
    u^{(n)}
    &=
    \epsilon^{\mathrm{ver}}-A_t^{(n)},
    \\
    \hat{v}_{\tau_i}^{(n)}
    &=
    v_\theta
    \left(
        x_{\tau_i}^{(n)},
        \tau_i
        \mid
        o_{t+d-1},
        l_t
    \right).
\end{aligned}
\label{eq:flowres_components}
\end{equation}
Here, $u^{(n)}$ is the flow direction implied by candidate
$A_t^{(n)}$, while $\hat{v}_{\tau_i}^{(n)}$ is the direction predicted
by the frozen policy for the same flow-time input under the pre-commit
observation. Their difference gives the velocity residual for candidate
$A_t^{(n)}$ at flow time $\tau_i$ under the pre-commit observation.

FlowRes aggregates these residuals over the selected action dimensions,
action positions, and verification flow times to obtain one score for
each candidate. Let $w_j\geq 0$ denote the weight assigned to action
position $j$, with $w_j=0$ for positions excluded from verification,
and let $\mathbf{m}$ mask the action dimensions included in
verification. For each verification flow time $\tau_i$, we compute
\begin{equation}
\begin{aligned}
    r_{\tau_i}^{(n)}
    &=
    \sum_{j=0}^{H-1}
    w_j
    \left\|
        \mathbf{m}
        \odot
        \left(
            \hat{v}_{\tau_i,j}^{(n)}
            -
            u_j^{(n)}
        \right)
    \right\|_2^2,
    \\
    E_n
    &=
    \sum_{i=1}^{M} r_{\tau_i}^{(n)},
    \\
    n^\star
    &=
    \arg\min_{1\leq n\leq N} E_n.
\end{aligned}
\label{eq:flowres_score}
\end{equation}

A lower $E_n$ indicates closer agreement between the candidate and the
frozen policy under the pre-commit observation. PACT-VLA uses this
policy-consistency score to rank the fixed candidate pool and selects
the candidate with the lowest $E_n$ for the next planned chunk switch.
FlowRes is therefore a policy-consistency score rather than a direct
estimate of task success. We evaluate whether this consistency is useful
for candidate selection in Section~\ref{sec:experiments}.

\subsection{Efficient Pipelined Execution}
\label{sec:efficient_execution}

PACT-VLA schedules candidate selection to finish before the planned
chunk switch at $t+d$. Since chunk switches occur only at control ticks,
meeting this deadline avoids adding another control-step delay.

Candidate generation begins at $t$ and uses shared computation and
batching where supported. At $t+d-1$, the controller reads the most
recent complete observation packet, which corresponds under our
control-step schedule to the pre-commit observation $o_{t+d-1}$.
For backbones with a reusable visual-language prefix, this observation
is processed immediately to produce a shared prefix KV cache and can
overlap in wall-clock time with the remaining candidate generation.
For flow policies without this module separation, the observation is
used directly in the later batched velocity query.

Let $t_r$ denote the earliest wall-clock instant at which both the
candidate pool and the conditioning required for verification are
ready. PACT-VLA then evaluates the $N\times M$ candidate and flow-time
inputs from Section~\ref{sec:flowres} in one batch and selects
$A_t^{(n^\star)}$ according to Eq.~\ref{eq:flowres_score}. Cached
backbones use a batched action-expert pass with the shared prefix cache;
other flow policies use a batched velocity query.

If selection finishes before $t+d$, the controller switches to
$A_t^{(n^\star)}$ at the originally scheduled control tick. Our latency
experiments evaluate this timing requirement on the evaluated hardware.\section{Experiments}
\label{sec:experiments}

We design our experiments and analyses to answer the following four questions.

\begin{enumerate}[leftmargin=2.0em]
    \item Does PACT-VLA improve task success over single-candidate execution
    and existing delay-robust methods across simulation benchmarks and
    real-robot tasks?

    \item Does FlowRes selection provide gains beyond simply generating more
    candidates, and how much of the gain comes from the observation used to
    score them?

    \item Does PACT-VLA transfer across inference delays, VLA backbones,
    benchmarks, and candidate-generation methods?

    \item Can PACT-VLA complete candidate selection before the planned chunk
    switch with practical latency?
\end{enumerate}

\subsection{Simulation Evaluation}
\label{sec:simulation}

We evaluate PACT-VLA on Kinetix~\citep{matthews2025kinetix} and
LIBERO~\citep{liu2023libero}. Kinetix tests fast control in dynamic and
contact-rich environments, while LIBERO evaluates image-based manipulation
with two large flow-based VLA models, $\pi_{0.5}$~\citep{intelligence2025pi_} and
GR00T N1.7~\citep{bjorck2025gr00t}.

\subsubsection{Kinetix}
\label{sec:kinetix}

\paragraph{Experimental Setup.}
Kinetix~\citep{matthews2025kinetix} contains 12 dynamic control tasks
involving behaviors such as throwing, catching, and balancing. Following
RTC~\citep{black2026real}, we use its released behavior-cloning checkpoint
with an action-chunk length of $H=8$. The controller replans every
$\max(1,d)$ control steps. All directly evaluated test-time methods use the
same checkpoint and execution protocol. PACT-VLA uses $N=4$ RTC-guided
candidates in the main comparison. We evaluate delays
$d\in\{0,1,2,3,4\}$ and report the average success rate across all 12 tasks
using 2,048 rollouts for each method and delay.

\paragraph{Compared Methods.}
We compare PACT-VLA with Naive Async, RTC~\citep{black2026real},
BID~\citep{liu2025bidirectional}, DreamChunk~\citep{chen2026dream},
PAINT~\citep{ho2026start}, and Training-time RTC~\citep{black2025training}.
Naive Async performs no delay correction. RTC and PAINT maintain
consistency with committed actions, BID selects among sampled chunks
using overlap with the previous chunk, DreamChunk uses later observations
to match predicted candidate trajectories, and Training-time RTC learns
delay handling during training. For RTC, we evaluate the soft-mask variant
and also include the reported hard-mask result.
\begin{table}[!ht]
\centering
\caption{
\textbf{Kinetix success rate (\%) across inference delays.}
Avg. denotes the mean over $d\in\{1,2,3,4\}$.
}
\label{tab:kinetix_main}
\footnotesize
\setlength{\tabcolsep}{6pt}
\begin{tabular*}{\linewidth}{@{\extracolsep{\fill}}lccccc}
\toprule
Method & $d{=}1$ & $d{=}2$ & $d{=}3$ & $d{=}4$ & Avg \\
\midrule
Naive asynchronous              & 75.1 & 71.1 & 58.7 & 48.9 & 63.5 \\
BID~\citep{liu2025bidirectional}                 & 87.9 & 81.2 & 68.7 & 55.8 & 73.4 \\
RTC~\citep{black2026real}                 & 89.3 & 84.8 & 75.0 & 62.2 & 77.8 \\
RTC, hard mask~\citep{black2026real} & 85.0 & 81.8 & 73.2 & 62.0 & 75.5 \\
PAINT-Euler~\citep{ho2026start}  & 85.0 & 80.4 & 73.2 & 63.8 & 75.6 \\
Training-time RTC~\citep{black2025training} & 88.5 & 86.0 & 80.0 & 72.1 & 81.7 \\
DreamChunk~\citep{chen2026dream}    & 87.0 & 82.5 & 70.8 & 58.5 & 74.7 \\
\textbf{PACT-VLA (ours)}        & \textbf{91.1} & \textbf{89.4} & \textbf{85.6} & \textbf{77.2} & \textbf{85.8} \\
\bottomrule
\end{tabular*}

\end{table}

\paragraph{Results.}
As shown in Table~\ref{tab:kinetix_main}, PACT-VLA achieves the highest
success rate at all four reported delays. Compared with RTC under the same
checkpoint and execution protocol, it increases average success from
$77.8\%$ to $85.8\%$. The gain over RTC grows from $1.8$ percentage points
at $d=1$ to $15.0$ points at $d=4$. At this longest delay, PACT-VLA reaches
$77.2\%$ success, compared with $62.2\%$ for RTC and $48.9\%$ for Naive
Async, showing stronger robustness as inference delay increases.

\begin{table}[!ht]
\centering
\caption{
\textbf{PACT-VLA combined with existing delay-robust methods on Kinetix.}
Each pair shows the success rates (\%) of a method with and without PACT-VLA.
}
\label{tab:compose}
\footnotesize
\setlength{\tabcolsep}{5pt}
\begin{tabular*}{\linewidth}{@{\extracolsep{\fill}}lccccc}
\toprule
Proposal / + selection & $d{=}1$ & $d{=}2$ & $d{=}3$ & $d{=}4$ & Avg \\
\midrule
VLASH~\citep{tang2025vlash} & 88.3 & 86.1 & 84.0 & 80.4 & 84.7 \\
\quad + \textbf{PACT-VLA}      & 89.4 & 88.6 & 87.3 & 85.0 & \textbf{87.5} \up{2.8} \\
\midrule
DEFLECT~\citep{zhu2026deflect}                  & 90.8 & 89.0 & 88.5 & 86.1 & 88.6 \\
\quad + \textbf{PACT-VLA}       & 91.3 & 90.3 & 90.1 & 88.7 & \textbf{90.1} \up{1.5} \\
\midrule
REMAC~\citep{wang2026real}          & 88.6 & 86.0 & 83.7 & 76.4 & 83.7 \\
\quad + \textbf{PACT-VLA}      & 89.8 & 87.1 & 85.3 & 78.4 & \textbf{85.2} \up{1.5} \\
\bottomrule
\end{tabular*}

\end{table}

Table~\ref{tab:compose} further shows that PACT-VLA improves
VLASH, DEFLECT, and REMAC at every evaluated delay. Average success
increases from $84.7\%$ to $87.5\%$ for VLASH, from $88.6\%$ to $90.1\%$
for DEFLECT, and from $83.7\%$ to $85.2\%$ for REMAC. Each method receives
its largest gain at $d=4$. These results show that PACT-VLA can select from
candidate pools produced by different delay-robust proposal methods and
further improve their performance.

\subsubsection{LIBERO}
\label{sec:libero}

\paragraph{Experimental Setup.}
We evaluate PACT-VLA on LIBERO~\citep{liu2023libero}, which tests
image-based manipulation across different spatial arrangements, objects,
and task goals. The evaluation covers the Spatial, Object, Goal, and
LIBERO-10 suites, with ten tasks in each suite.

We evaluate two independently developed VLAs,
$\pi_{0.5}$~\citep{intelligence2025pi_} and
GR00T N1.7~\citep{bjorck2025gr00t}. For each backbone, the baseline is the
corresponding asynchronous policy that generates a single candidate chunk.
PACT-VLA uses the same policy, checkpoint, and execution protocol, but
generates $N=8$ candidates and applies FlowRes selection before commitment.
We follow the standard evaluation pipeline of each backbone and evaluate
delays $d\in\{1,2,3,4\}$. The controller executes the first five actions
of each chunk for $\pi_{0.5}$ and the first eight for GR00T N1.7 before
switching to the next chunk. Within each backbone, the baseline and PACT-VLA
use the same execution horizon. Each method is evaluated with 500 episodes
per suite and delay.

\begin{table}[!ht]
\centering
\caption{
\textbf{LIBERO success rate (\%) for $\pi_{0.5}$ and GR00T N1.7.}
Each block compares single-candidate asynchronous execution with PACT-VLA.
}
\label{tab:libero}
\footnotesize
\setlength{\tabcolsep}{2pt}
\begin{tabular*}{\linewidth}{@{\extracolsep{\fill}}l cc cc cc cc cc}
\toprule
 & \multicolumn{2}{c}{Goal} & \multicolumn{2}{c}{Spatial} & \multicolumn{2}{c}{Object} & \multicolumn{2}{c}{LIBERO-10} & \multicolumn{2}{c}{Average} \\
\cmidrule(lr){2-3}\cmidrule(lr){4-5}\cmidrule(lr){6-7}\cmidrule(lr){8-9}\cmidrule(lr){10-11}
Delay & Single & PACT-VLA & Single & PACT-VLA & Single & PACT-VLA & Single & PACT-VLA & Single & PACT-VLA \\
\midrule
\multicolumn{11}{c}{\emph{$\pi_{0.5}$ }} \\
\midrule
\dg{$d{=}1$} & \dg{92.8} & \dg{94.6} & \dg{96.0} & \dg{94.2} & \dg{97.0} & \dg{96.4} & \dg{91.2} & \dg{90.4} & \dg{94.2} & \dg{93.9} \dn{0.3} \\
$d{=}2$ & 91.4 & 91.6 & 85.6 & 86.4 & 93.8 & 95.0 & 80.2 & 86.0 & 87.8 & 89.8 \up{2.0} \\
$d{=}3$ & 80.2 & 85.8 & 59.0 & 64.8 & 89.0 & 90.0 & 77.0 & 78.2 & 76.3 & 79.7 \up{3.4} \\
$d{=}4$ & 68.2 & 70.0 & 43.8 & 47.8 & 65.0 & 70.8 & 62.8 & 65.2 & 60.0 & 63.5 \up{3.5} \\
\midrule
\multicolumn{11}{c}{\emph{GR00T N1.7}} \\
\midrule
\dg{$d{=}1$} & \dg{93.6} & \dg{93.2} & \dg{96.0} & \dg{96.2} & \dg{96.4} & \dg{96.6} & \dg{87.6} & \dg{85.2} & \dg{93.4} & \dg{92.8} \dn{0.6} \\
$d{=}2$ & 86.2 & 89.2 & 88.0 & 89.6 & 92.2 & 93.4 & 72.6 & 76.2 & 84.8 & 87.1 \up{2.3} \\
$d{=}3$ & 67.6 & 73.0 & 73.0 & 75.0 & 82.2 & 87.0 & 55.4 & 62.0 & 69.6 & 74.3 \up{4.7} \\
$d{=}4$ & 52.2 & 58.2 & 53.6 & 56.8 & 62.6 & 67.8 & 36.4 & 42.6 & 51.2 & 56.4 \up{5.2} \\
\bottomrule
\end{tabular*}
\end{table}

\paragraph{Results.}
Table~\ref{tab:libero} reveals a clear delay-dependent pattern. At $d=1$, verification sees the same observation as generation,
not an updated view of the scene. PACT-VLA can still select a
different chunk, but a lower score under the old observation does
not necessarily mean better execution. This selection can help or
hurt task success. On LIBERO, average success decreases slightly
on both backbones in this setting. At $d=2,3,4$, the pre-commit observation is more
recent than the generation observation, and both backbones improve on
average. Across the tested delays, the average success-rate gain over
single-candidate execution increases with inference delay on both
backbones. At $d=4$, the gain reaches $3.5$ percentage points on
$\pi_{0.5}$ and $5.2$ points on GR00T N1.7.

At $d=4$, PACT-VLA improves all four LIBERO suites on both backbones.
On GR00T N1.7 at this delay, its average success is 4.2 percentage
points higher than random selection and 3.8 points higher than
FlowRes scoring under the generation observation.
All three selectors use $N=8$ candidates
(Appendix~\ref{app:selector_ablations}).

\subsection{Real-World Evaluation}
\label{sec:real_world}

We deploy an in-house $\pi_{0.5}$~\citep{intelligence2025pi_} policy on a bimanual AgileX PiperX
platform with two 6-DoF arms. The robot is equipped with three Intel
RealSense D435i cameras, including one overhead camera and one wrist
camera on each arm. Our system is built on
LeRobot~\citep{cadene2024lerobot}.

\subsubsection{Task Success}
\label{sec:real_world_tasks}

\paragraph{Experimental Setup.}
We use an action-chunk length of $H=50$ and 10 denoising steps.
The robot operates at 30 Hz, and the asynchronous controller requests a
new prediction when half of the current action queue remains.
Using $\pi_{0.5}$, we compare Naive Async, VLASH, Old-FlowRes, and
PACT-VLA. Naive Async, Old-FlowRes, and PACT-VLA use the same frozen
task-specific $\pi_{0.5}$ checkpoint obtained by standard SFT.
PACT-VLA requires no further training. Naive Async generates a single
action chunk, while Old-FlowRes and PACT-VLA each generate $N=4$
candidates. Old-FlowRes scores them using the observation available at
generation start, whereas PACT-VLA uses the pre-commit observation.
VLASH is evaluated separately using its own checkpoint.

We evaluate three everyday manipulation tasks.
\textbf{Toasting Bread} requires placing two slices into a toaster and
retrieving them afterward.
\textbf{Loading Dishwasher} requires loading bowls and a plate into a
dishwasher and later returning them to the table.
\textbf{Pouring Coffee Beans} requires scooping beans onto a plate and
pouring them back into the container.
For each task, we conduct 50 trials per method and report full-task
success, which requires all task sub-goals to be completed.

\begin{figure}[t]
\centering
\includegraphics[width=0.96\linewidth]{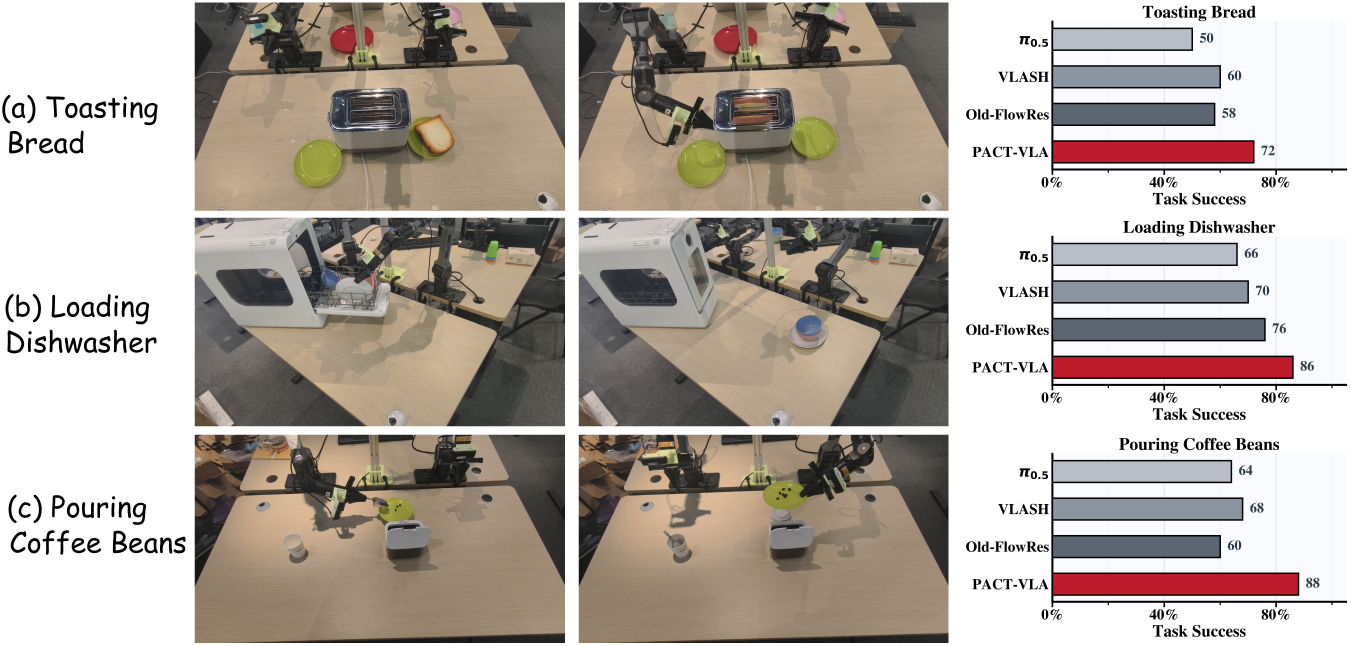}
\caption{
\textbf{Real-world evaluation.} Example task stages and full-task success rates (\%) for $\pi_{0.5}$ with Naive Async execution, VLASH, Old-FlowRes, and PACT-VLA. Old-FlowRes scores candidates using the generation observation, while PACT-VLA uses the pre-commit observation.
}
\label{fig:realrobot}
\end{figure}

\paragraph{Results.}
Figure~\ref{fig:realrobot} shows that PACT-VLA achieves the highest success
rate on all three tasks. Average success is $60.0\%$ for Naive Async,
$66.0\%$ for VLASH, $64.7\%$ for Old-FlowRes, and $82.0\%$ for PACT-VLA.
Compared with Old-FlowRes, PACT-VLA improves success by 14 percentage
points on Toasting Bread, 10 points on Loading Dishwasher, and 28 points
on Pouring Coffee Beans, for an average gain of 17.3 points. Since the
policy checkpoint, candidate budget, and execution rule are unchanged,
this comparison supports using the pre-commit observation for candidate
selection. PACT-VLA also achieves 16.0 percentage points higher average
success than VLASH in this evaluation.

\subsubsection{Deployment Latency}
\label{sec:deployment_latency}

{\setlength{\intextsep}{3pt}
\begin{table}[!ht]
\centering
\setlength{\abovecaptionskip}{8pt}
\caption{
\textbf{Generation and verification latency across GPUs.}
Each entry is the P99 latency measured on that GPU.
The control-step delay $d$ is calculated from the listed latency at 30 Hz.
}
\label{tab:deployment_latency}
\small
\setlength{\tabcolsep}{6pt}
\begin{tabular*}{\linewidth}{@{\extracolsep{\fill}}lcccc}
\toprule
\raisebox{-7pt}[0pt][0pt]{GPU} & \multicolumn{2}{c}{Generation ($N=1$)}
& \multicolumn{2}{c}{Generation + Verification ($N=4$)} \\
\cmidrule(lr){2-3}
\cmidrule(lr){4-5}
& Latency (ms) & $d$ & Latency (ms) & $d$ \\
\midrule
RTX 5090 & 35.4  & 2 & 60.2  & 2 \\
H100     & 42.5  & 2 & 58.8  & 2 \\
RTX Pro 6000 & 48.7  & 2 & 64.8  & 2 \\
\bottomrule
\end{tabular*}
\end{table}
\par}

Table~\ref{tab:deployment_latency} compares single-candidate generation
with four-candidate generation and FlowRes verification across three GPUs.
Appendix~\ref{app:latency_breakdown} resolves the RTX 5090 figure into the stages of one
decision cycle.

At 30 Hz, the controller can switch action chunks only every $33.33$ ms.
On all three GPUs, both configurations take more than one
control period but less than two, corresponding to $d=2$.
Thus, although PACT-VLA requires additional
computation, the reported runtimes correspond to the same control-step
delay as single-candidate generation on each GPU.\section{Conclusion}

We presented PACT-VLA, a training-free framework that uses inference delay
as decision time for action-chunk selection. While candidate generation is
underway, observations arriving during inference can guide the pre-commit
choice of one complete action chunk, with FlowRes reusing the frozen flow
policy to rank the candidates. Across Kinetix, LIBERO, and three real-robot
tasks, PACT-VLA improves delayed execution, complements existing delay-robust
methods, and raises real-robot average success from 60\% to 82\% using the
same checkpoint.
\section*{AI use statement}

We used generative AI tools (Claude Code, based on Anthropic's Claude models)
throughout this project. In the categories for which ICLR requires disclosure,
we used these tools to implement methods, to design and to give feedback on our
experimental methodology, to propose and refine hypotheses, to reformat and
aggregate evaluation logs, and to interpret results. We did not use generative
AI to generate synthetic datasets, to develop theoretical models or conceptual
frameworks, or to formulate, support, or write proofs of mathematical claims;
these categories are not applicable to this work. In the categories for which
disclosure is recommended, we additionally used these tools to write and edit
software code, to draft and to edit parts of this paper, to identify and summarise related literature, and to format
references.

We have reviewed all AI-assisted work, and we validated the AI-assisted
evaluation code against external reference points rather than by inspection
alone. Our runs of published baselines are checked against the numbers their
authors report, to the precision those numbers can be read off published
figures, and an $N{=}1$ control recovers the RTC row through our own execution
path (Appendix~\ref{app:experimental_settings}); the released
$\pi_{0.5}$ LIBERO checkpoint reproduces its published clean success rate on
our harness before any of our own machinery is enabled; and where a cell was
run under two independent seeds we report their agreement. Implementation
errors found this way, including one in an early scoring implementation
that silently regressed onto zero-padded action dimensions and one in which
replicated candidates aliased a single mutable input, were fixed and every
affected result was recomputed.

We take responsibility for the final content of this work, including all text,
claims, and artifacts produced with the aid of generative AI.

\bibliography{references}
\bibliographystyle{iclr2027_conference}

\appendix
\section{Related Work}
\label{sec:related_work}

\paragraph{Action-Chunk Execution under Inference Delay.}
Existing methods address delayed execution by improving chunk continuity
or adapting the policy to the expected execution state.
RTC~\citep{black2026real} uses inpainting guidance to align new predictions with
committed actions, while PAINT~\citep{ho2026start} achieves prefix consistency
through initial noise selection. VLASH~\citep{tang2025vlash} uses future-state
conditioning, and Training-time RTC~\citep{black2025training} learns to generate
chunks from committed action prefixes. REMAC~\citep{wang2026real} learns
corrective adjustments through masked action chunking.
DEFLECT~\citep{zhu2026deflect} instead constructs offline preferences from
future and stale observations to improve the policy under its deployment
input. These approaches primarily change how action chunks are generated.
PACT-VLA complements them by keeping their generation procedure and using
the pre-commit observation to select among the resulting candidates.

\paragraph{Test-Time Action Selection and Verification.}
Generating multiple candidates allows a policy to compare alternatives
rather than execute its first prediction. Existing approaches differ in
the information used for this comparison. BID~\citep{liu2025bidirectional} uses backward
coherence and forward contrast to guide candidate selection.
RoboMonkey~\citep{kwok2025robomonkey} employs a learned VLM verifier, whereas
MG-Select~\citep{jang2026verifier} derives confidence from the policy's
action-token distributions.
$\tau_0$-WM~\citep{zhou2026tau_0} uses re-denoising consistency to rank
candidates under the generation observation. PACT-VLA uses a newer
pre-commit observation to verify previously generated candidates, with
selection completed before the planned chunk switch. DreamChunk~\citep{chen2026dream} also incorporates
observations collected after generation. It predicts candidate latent
trajectories with a learned world model and matches them against the
observed rollout. PACT-VLA instead scores the existing chunks directly
with the frozen flow policy under the pre-commit observation.
Its distinction is therefore the combination of updated observations
and policy-based candidate verification, without requiring a separate
learned verifier or predictive model.\section{Experimental Settings and Statistical Reporting}
\label{app:experimental_settings}

\paragraph{Evaluation settings.}
Table~\ref{tab:experiment_settings} summarizes the main simulation configurations,
including the execution protocol, checkpoint, candidate-generation method, candidate count,
evaluation count, and execution-noise level.
The LIBERO selector comparisons in Appendix~\ref{app:selector_ablations} use the GR00T
configuration in the last row, with $N{=}1$ for the single-candidate baseline and $N{=}8$
for the three multi-candidate methods.

\begin{table}[!ht]
\centering\scriptsize
\caption{What differs between experiments. ``Proposal'' is where the $N$ candidates come
from. The main PACT-VLA configuration verifies candidates under the pre-commit observation
$o_{t+d-1}$; alternative selectors and verification observations are specified in
Appendix~\ref{app:selector_ablations}. Evaluation counts are per Kinetix level or LIBERO
suite for each method and delay.}
\label{tab:experiment_settings}
\setlength{\tabcolsep}{1.6pt}
\begin{tabular*}{\linewidth}{@{\extracolsep{\fill}}llllllc}
\toprule
Result & Protocol & Ckpt & Proposal & $N$ & Evaluation count & $\sigma$ \\
\midrule
Table~\ref{tab:kinetix_main} & continuous, $H{=}8$, $d{\leq}4$ & \texttt{bc/31} & RTC-guided & 4 & 2048 & 0.1 \\
Table~\ref{tab:compose}, VLASH & continuous, $H{=}8$, $d{\leq}4$ & async5 & speculative & 1 / 4 & 2048 & 0.1 \\
Table~\ref{tab:compose}, DEFLECT & continuous, $H{=}8$, $d{\leq}4$ & DEFLECT & speculative & 1 / 4 & 2048 & 0.1 \\
Table~\ref{tab:compose}, REMAC & continuous, $H{=}8$, $d{\leq}4$ & REMAC LoRA & warm-start refine & 1 / 4 & 2048 & 0.1 \\
Table~\ref{tab:libero}, $\pi_{0.5}$ & predict 50, execute 5, $d{\leq}4$ & $\pi_{0.5}$ LIBERO & policy sampling & 1 / 8 & 500 & 0 \\
Table~\ref{tab:libero}, GR00T & predict 16, execute 8, $d{\leq}4$ & GR00T N1.7 per suite & policy sampling & 1 / 8 & 500 & 0 \\
\bottomrule
\end{tabular*}
\end{table}

\paragraph{FlowRes configuration.}
Table~\ref{tab:flowres_settings} gives the values the reported runs use. All three share one
noise sample across the candidates and all verification times, and all three score under
$o_{t+d-1}$ and commit at $t+d$.

\begin{table}[h]
\centering\footnotesize
\caption{FlowRes settings for the simulation experiments. ``Scored steps'' is the range of $j$ entering the sum,
``weights'' the $w_j$ over that range, and ``effective dims'' the entries kept by $\mathbf{m}$.}
\label{tab:flowres_settings}
\setlength{\tabcolsep}{5pt}
\begin{tabular*}{\linewidth}{@{\extracolsep{\fill}}lllll}
\toprule
Model & $\mathcal{T}$ & Scored steps & $w_j$ & Effective dims \\
\midrule
Kinetix flow policy & $0.2,0.4,0.6,0.8$ & all 8 predicted & $0.5^{\,j}$ & all \\
$\pi_{0.5}$ (LIBERO) & $0.2,0.4,0.6,0.8$ & first 5 (the executed ones) & uniform & 7 of 32 \\
GR00T N1.7 & $0.1,0.3,0.5,0.7$ & all 16 predicted & $0.5^{\,j}$ & 7 of 132 \\
\bottomrule
\end{tabular*}
\end{table}

\paragraph{Baseline implementation check.}
For the main Kinetix configuration in Table~\ref{tab:kinetix_main}, setting $N{=}1$ uses
RTC-guided generation without candidate selection.
The resulting success rates differ from the RTC baseline by at most
$0.7$ percentage points across the reported delays.

\paragraph{Statistical reporting.}
For Table~\ref{tab:kinetix_main}, each method is evaluated with $2{,}048$ rollouts per
Kinetix level and delay, giving $24{,}576$ rollouts across the twelve levels.
Table~\ref{tab:compose} uses the counts listed in Table~\ref{tab:experiment_settings}.
For Table~\ref{tab:libero}, each method is evaluated with $500$ episodes per LIBERO suite
and delay. Benchmark averages are unweighted means over levels or suites.
Full-rollout comparisons use separate runs for each method.
The fixed-pool evaluation in Appendix~\ref{app:fixed_pool} instead compares selection
rules at the same saved decision points.
We report observed success rates and their differences without confidence intervals.

\section{Deployment Latency Breakdown}
\label{app:latency_breakdown}

\begin{figure}[t]
\centering
\includegraphics[width=0.72\linewidth]{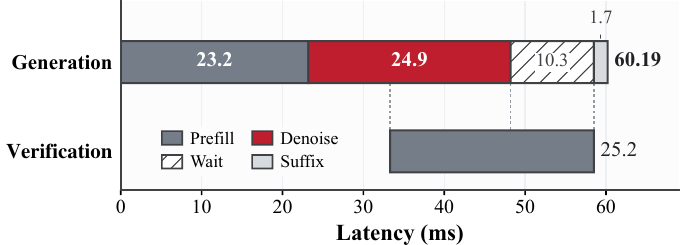}
\caption{
One decision cycle on an RTX 5090 with $\pi_{0.5}$ and $N{=}4$.
Generation and verification share the same GPU.
Encoding the pre-commit observation overlaps with candidate generation.
Once the candidates are ready, scoring waits for this encoding to
finish and then takes $1.7$ ms. The full cycle takes $60.19$ ms.
}
\label{fig:latency_breakdown}
\end{figure}

Figure~\ref{fig:latency_breakdown} resolves the RTX 5090 entry of
Table~\ref{tab:deployment_latency} into the stages of one decision cycle.

Generation begins at inference start with $23.2$ ms of prefill over $o_t$, followed by
$24.9$ ms of flow integration. Generation and verification run concurrently on the same GPU.
The pre-commit observation arrives one control step in, at $33.33$ ms, and its prefill
occupies the next $25.2$ ms. Because that prefill depends only on the observation and not on
the candidates, it overlaps the tail of generation instead of following it.

Candidate scoring starts only after both the candidates and the shared
prefix KV cache are ready. The candidates become available first and
wait about $10.3$ ms for the cache before the $1.7$ ms scoring pass.
The full cycle ends at $60.19$ ms. The time after candidate generation
therefore includes both waiting and scoring. The segments are not independent: verification prefill shares
the GPU with the tail of generation, so the generation time measured here already includes
that contention and the figure does not separate the cost of producing four candidates from
the cost of running verification alongside them. At 30 Hz the whole cycle still fits inside two control periods, which is the $d{=}2$
reported for both configurations on this GPU.

\section{Selector Ablations}
\label{app:selector_ablations}

\paragraph{Experimental controls.}
Tables~\ref{tab:selector_comparison} and~\ref{tab:verification_delay} use the GR00T
configuration of Table~\ref{tab:libero}, including the same per-suite checkpoints, execution
protocol, and episode count.
The single-candidate baseline generates and executes one chunk.
The three multi-candidate methods each generate $N{=}8$ candidates.
Random selection chooses one uniformly at random.
FlowRes on $o_t$ scores candidates using the generation observation,
while PACT-VLA uses the pre-commit observation $o_{t+d-1}$.

All methods use the same task set, initial-state set, inference delay,
eight-action execution horizon, episode limit, and chunk-switching
rule. The three multi-candidate methods also use the same generation
procedure. Each method runs separately, and the candidate-sampling
noise is not explicitly seeded. Candidates are generated from the
observations encountered in each method's own rollout, which may
differ after different actions are executed. These experiments
therefore compare full-rollout success rather than selection from
one shared candidate pool. The baseline column of
Table~\ref{tab:selector_comparison} is the GR00T run reported in
Table~\ref{tab:libero}.

\paragraph{Selector comparison.}
At $d{=}4$, average success is $51.2\%$ for single-candidate execution, $52.2\%$ for random
selection, $52.6\%$ for FlowRes under the generation observation, and $56.4\%$ for PACT-VLA.
Scoring under the pre-commit observation yields higher observed success than
same-observation scoring on all four suites.

\begin{table}[h]
\centering\footnotesize
\caption{
Selector comparison on GR00T N1.7 at $d{=}4$, with success rates
reported in percent. All methods use the same evaluation protocol
and initial-state set. The baseline generates one candidate,
while the three multi-candidate methods each generate eight.
Each method runs separately without sharing candidate pools.
}
\label{tab:selector_comparison}
\setlength{\tabcolsep}{6pt}
\begin{tabular*}{\linewidth}{@{\extracolsep{\fill}}lcccc}
\toprule
Suite & Baseline & Random pick & FlowRes on $o_t$ & PACT-VLA \\
\midrule
Goal      & 52.2 & 55.8 & 54.2 & 58.2 \\
Spatial   & 53.6 & 53.2 & 52.8 & 56.8 \\
Object    & 62.6 & 64.8 & 64.6 & 67.8 \\
LIBERO-10 & 36.4 & 35.0 & 38.6 & 42.6 \\
\midrule
Average   & 51.2 & 52.2 & 52.6 & 56.4 \\
\bottomrule
\end{tabular*}
\end{table}

\paragraph{Effect of the verification observation across delays.}
Table~\ref{tab:verification_delay} compares FlowRes under the generation and pre-commit
observations. The observed average difference is larger at $d{=}3$ and $d{=}4$ than at
$d{=}2$. At $d{=}1$, both variants use the same observation; the reported values come from
independent runs.

\begin{table}[h]
\centering\footnotesize
\caption{\textbf{PACT-VLA minus FlowRes on $o_t$ for GR00T N1.7}, percentage points, by
inference delay. At $d{=}1$ both arms score under the same observation.}
\label{tab:verification_delay}
\setlength{\tabcolsep}{8pt}
\begin{tabular*}{\linewidth}{@{\extracolsep{\fill}}l>{\color{black!45}}cccc}
\toprule
Suite & $d{=}1$ & $d{=}2$ & $d{=}3$ & $d{=}4$ \\
\midrule
Goal      & $-2.8$ & $+1.2$ & $+3.0$ & $+4.0$ \\
Spatial   & $+1.0$ & $+1.8$ & $+5.0$ & $+4.0$ \\
Object    & $+0.4$ & $+0.4$ & $+2.6$ & $+3.2$ \\
LIBERO-10 & $-0.6$ & $+0.8$ & $+6.8$ & $+4.0$ \\
\midrule
Average   & $-0.5$ & $+1.1$ & $+4.3$ & $+3.8$ \\
\bottomrule
\end{tabular*}
\end{table}

\paragraph{Interpretation and scope.}
At $d{=}1$ the pre-commit observation coincides with the generation observation, and
selection can help or hurt: per-suite differences in Table~\ref{tab:libero} span $-1.8$ to
$+1.8$ on $\pi_{0.5}$ and $-2.4$ to $+0.2$ on GR00T N1.7. Under the five-step execution
horizon of that table, the $\pi_{0.5}$ Spatial and Object suites came out negative at
$d{=}1$ in all three evaluation seeds we ran, by $-1.2$ and $-0.7$ points on average.

\subsection{Observation updates on Kinetix}
\label{app:kinetix_observation}

\paragraph{Setting.}
We compare two FlowRes selectors on 12 Kinetix levels. Both use $N=4$ candidates, the same checkpoint,
candidate-generation procedure, scoring parameters, and execution
schedule. They differ only in the observation used for verification:
Old scores under the generation observation $o_t$, whereas PACT-VLA
scores under the pre-commit observation $o_{t+d-1}$.
We vary the inference delay $d$ and execution-noise level $\sigma$.
Table~\ref{tab:kinetix_observation} reports episode success averaged
over levels.

\begin{table}[t]
    \centering
    \caption{
        Episode success (\%) on Kinetix when changing only the
        observation used for FlowRes scoring.
        Results are averaged over 12 levels.
    }
    \label{tab:kinetix_observation}
    \begin{tabular*}{\linewidth}{@{\extracolsep{\fill}}ccrr}
        \hline
        $d$ & $\sigma$ & Old ($o_t$) & PACT-VLA ($o_{t+d-1}$) \\
        \hline
        2 & 0.1 & 85.44 & 89.40 \\
        3 & 0.1 & 75.04 & 85.56 \\
        4 & 0.1 & 62.02 & 77.16 \\
        \hline
        4 & 0.0 & 64.69 & 79.64 \\
        4 & 0.2 & 52.07 & 64.35 \\
        \hline
    \end{tabular*}
\end{table}

\paragraph{Results.}
At $\sigma=0.1$, the difference grows with the inference delay, from
3.96 percentage points at $d=2$ to 15.14 percentage points at $d=4$.
The gain remains positive without injected execution noise
(14.95 points at $\sigma=0$) and under stronger noise
(12.28 points at $\sigma=0.2$).
These results support updating the scoring observation under inference delay.
The benefit does not require injected execution noise.

\subsection{Fixed-pool branch evaluation}
\label{app:fixed_pool}

\paragraph{Setting.}
At $d=4$ and $\sigma=0.1$, we evaluate 1,024 saved decision points
per level and seed across 12 Kinetix levels and three seeds.
Each point has a fixed pool of $N=4$ candidates.
Old and PACT-VLA select their candidates before any branch rollout.
Each candidate is executed for four steps from the same saved state,
after which all branches use the same continuation controller
until termination. We perform eight independent continuations
per candidate and evaluate uniform random selection, Old, and
PACT-VLA on the same candidate pools.
Results include all saved decision points.

\paragraph{Results.}
Average branch success is 61.18\% for random selection,
61.07\% for Old, and 61.62\% for PACT-VLA.
PACT-VLA improves over Old by 0.56 percentage points, with gains
of 0.55, 0.49, and 0.63 points across the three seeds.
Old remains close to random selection in observed success, while
the pre-commit observation improves selection on unchanged
candidate pools.
This comparison isolates a single selection decision under a
common continuation controller.
Its gains are not directly comparable to episode-level gains
from repeatedly applying the selector throughout a rollout.

\section{Real-Robot Data and Training}
\label{app:real_robot_training}

\paragraph{Demonstration data.}
We collect task-specific demonstrations through leader--follower teleoperation on the platform
described in Section~\ref{sec:real_world}. Each demonstration records three camera streams at
$480 \times 640$ resolution, together with the 14-dimensional robot state and action, at
30\,Hz. Each task uses only its own demonstrations.
Table~\ref{tab:real_robot_data} summarizes the datasets, which contain 675 demonstrations
totaling approximately 9.19 hours.

\begin{table}[h]
    \centering
    \caption{Task-specific real-robot demonstration datasets.}
    \label{tab:real_robot_data}
    \begin{tabular*}{\linewidth}{@{\extracolsep{\fill}}lrrr}
        \toprule
        Task & Demonstrations & Duration (h) & Mean length (s) \\
        \midrule
        Toasting Bread       & 349 & 3.24 & 33.4  \\
        Loading Dishwasher   & 110 & 3.41 & 111.7 \\
        Pouring Coffee Beans & 216 & 2.54 & 42.4  \\
        \midrule
        Total                & 675 & 9.19 & \\
        \bottomrule
    \end{tabular*}
\end{table}

\paragraph{Base-policy training.}
We independently fine-tune a $\pi_{0.5}$ policy for each task using standard SFT for 40,000
gradient steps and evaluate the final checkpoint. Training uses AdamW with a batch size of 8,
a peak learning rate of $10^{-5}$, a 1,000-step linear warmup, and cosine decay to 1\% of the
peak rate. Only the action expert is updated; the vision-language backbone remains frozen.
Naive Async, Old-FlowRes, and PACT-VLA use the same frozen task-specific SFT checkpoint. None of these three methods uses delay-aware or preference post-training. On the robot, Old-FlowRes and PACT-VLA each generate $N=4$ candidates
and use the verification flow times listed for $\pi_{0.5}$ in
Table~\ref{tab:flowres_settings}.
Both methods score the first 25 actions with uniform weights, and
the mask $\mathbf{m}$ retains all 14 bimanual action dimensions.
This 25-action window determines which actions contribute to FlowRes,
not when the controller switches chunks.
Old-FlowRes uses the generation observation for scoring, while
PACT-VLA uses the pre-commit observation.
The policy predicts $H=50$ actions, and the controller requests a
new prediction when half of the current action queue remains,
as described in Section~\ref{sec:real_world}.

\end{document}